\def\year{2022}\relax
\documentclass[letterpaper]{article} 
\usepackage{aaai22}  
\usepackage{times}  
\usepackage{helvet}  
\usepackage{courier}  
\usepackage[hyphens]{url}  
\usepackage{graphicx} 
\usepackage{natbib}  
\usepackage{caption} 
\DeclareCaptionStyle{ruled}{labelfont=normalfont,labelsep=colon,strut=off} 
\usepackage[switch]{lineno} 
\usepackage{amssymb}
\usepackage{amsmath}
\usepackage{float}
\usepackage{tabularx}
\usepackage{caption}
\usepackage[ruled,vlined]{algorithm2e}
\usepackage{amsmath}
\usepackage{booktabs}
\usepackage{footnote}
\usepackage{tablefootnote}
\author{Jiashu Pu, Guandan Chen, Yongzhu Chang, XiaoxiMao\thanks{corresponding author}\\
  {\normalsize Fuxi AI Lab, NetEase Inc., Hangzhou, China} \\
  \texttt{\normalsize \{pujiashu,chenguandan,changyongzhu,maoxiaoxi\}@corp.netease.com}}
\title{Dialog Intent Induction via Density-based Deep Clustering Ensemble}
\usepackage{bibentry}

\begin{document}

\maketitle

\begin{abstract}
Existing task-oriented chatbots heavily rely on spoken language understanding~(SLU) systems to determine a user's utterance's intent and other key information for fulfilling specific tasks. In real-life applications, it is crucial to occasionally induce novel dialog intents from the conversation logs to improve the user experience. In this paper, we propose the \textbf{D}ensity-based \textbf{D}eep \textbf{C}lustering \textbf{E}nsemble~(DDCE) method for dialog intent induction. Compared to existing K-means based methods, our proposed method is more effective in dealing with real-life scenarios where a large number of outliers exist. To maximize data utilization, we jointly optimize texts' representations and the hyperparameters of the clustering algorithm. In addition, we design an outlier-aware clustering ensemble framework to handle the overfitting issue. Experimental results over seven datasets show that our proposed method significantly outperforms other state-of-the-art baselines.
\end{abstract}

\section{Introduction}
\label{sec:intro}

In recent years, applications built with task-oriented chatbots have become ubiquitous in many fields~\cite{perkins2019dialog}. Despite the considerable success that massive pre-training has achieved over open-domain response generation, task-oriented chatbots still heavily rely on SLU systems to convert a user's utterance to a specified dialog intent and corresponding slots information. In general, the SLU system is trained on a handful of examples corresponding to several pre-determined dialog intents before deployment. However, in real-life scenarios, it's common that dialog intents designed by developers may not cover actual users' utterances. User demands may shift with time or are simply not considered in advance. To perfect the user experience, developers need to induce novel dialog intents from the conversation logs generated by users. 

In practice, due to the enormous size of conversation logs, developers typically perform a clustering analysis over the logs to find clusters with a large number of similar user utterances and then mark them as novel dialog intents. Because of the practicability of this task, there have been multiple works proposed by researchers. However, these works are mostly based on K-means algorithms~\cite{hadifar2019self,perkins2019dialog,lin2020discovering,wang2016semi}. In practice, these K-means based methods have two limitations. First, the hyperparameter $K$ is challenging to determine before clustering. Choosing a proper $K$ takes a lot of trial and error. Second, there are a large number of outliers in real-world conversation logs. These outliers are occasional, irrelevant user utterances that should not be mapped to any dialog intent. They cannot be effectively excluded by K-means based methods and take much human labor to clean. To demonstrate, we present an example in Figure~\ref{fig:DDEC_intent_cluster}.

\begin{figure}[t]
    \centering
\includegraphics[width=0.45\textwidth]{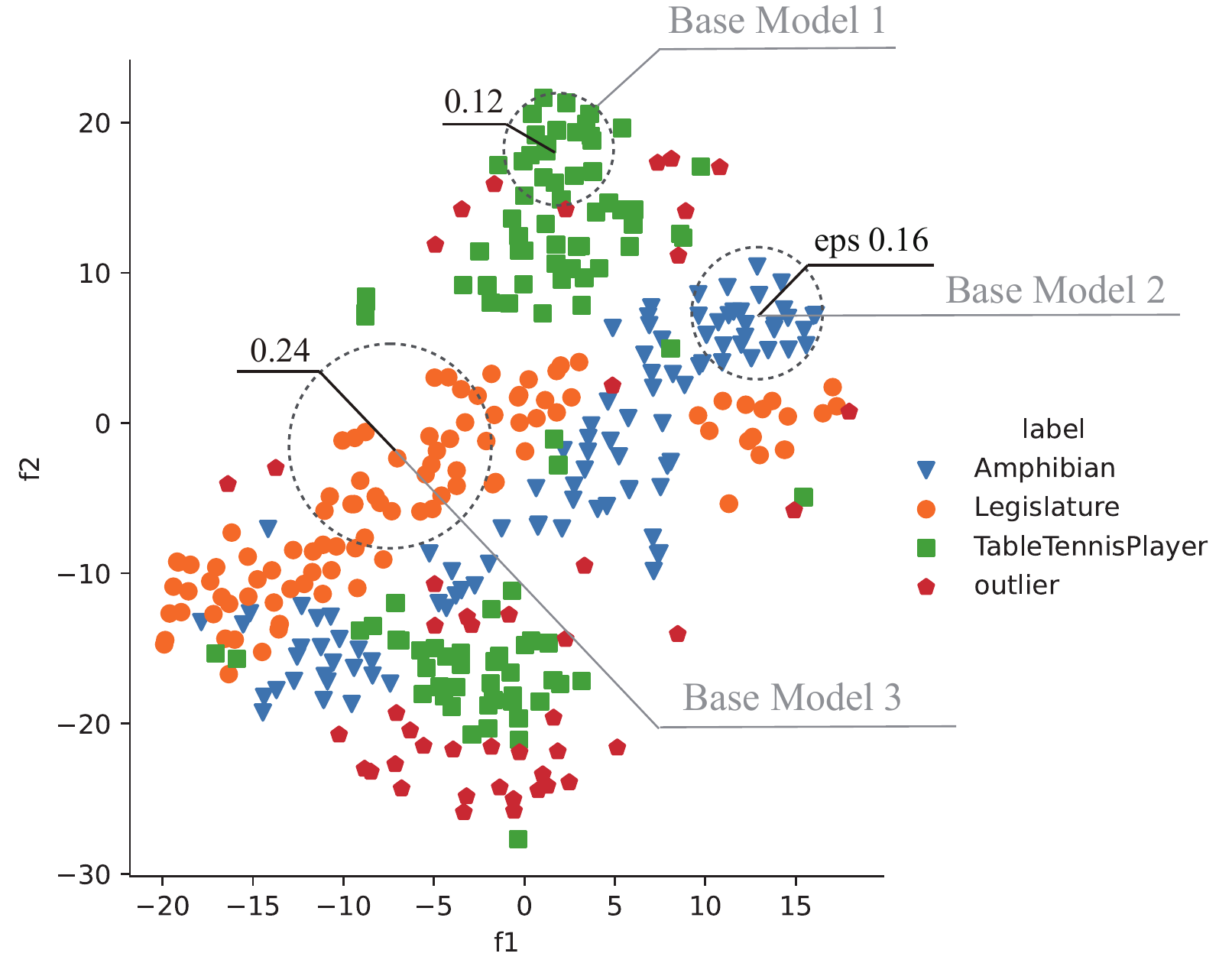}
    \caption{
    We apply t-SNE on embeddings of a small fraction of samples from DBpedia~\cite{auer2007dbpedia}. The embeddings are computed with BERT-base~\cite{devlin2019bert} without finetuning. It is difficult to distinguish labels from the density of aggregated samples, and the involvement of outliers makes the task even more difficult. Besides, the figure illustrates a clustering model with a specific set hyperparameters is effective on only a part of the data ($eps$ refers to the radius of neighborhood defined in DBSCAN~\cite{ester1996density}).}
    \label{figure1}
\end{figure}

In this paper, we propose a method called \textbf{D}ensity based \textbf{D}eep \textbf{C}lustering \textbf{E}nsemble(DDCE). In this method, we adopt a density-based clustering algorithm OPTICS~\cite{ankerst1999optics}, to avoid the limitations mentioned above. In addition, it's widely known that general representations obtained with pre-trained language models are not sufficient to support effective clustering~\cite{li2020sentence,reimers2019sentence}, as such representations sometimes may fail to distinguish between specific semantics (e.g. negation), which is illustrated in Figure~\ref{fig:DDEC_intent_cluster} and Figure~\ref{figure1}; besides, in practice, we also notice that a specific set of hyperparameters of a clustering algorithm 
may not be effective across the whole data.

Due to this, we propose to apply a clustering ensemble framework that combines multiple base clustering models with corresponding text encoders and hyperparameters. Cluster ensemble has been proven to enhance the robustness and thus improve clustering quality~\cite{strehl2002cluster}. Although the concept of ensemble learning has been well established for tasks such as classification and regression~\cite{strehl2002cluster,bauer1999empirical}, there is very little work that applies it to the clustering problem of novel intent induction. Our framework that combines representation learning and clustering ensemble fills this void. We conduct detailed experiments on seven datasets to verify its effectiveness. The experimental results show that our method significantly outperforms other state-of-the-art baselines.

\begin{figure*}[th]
    \centering
\includegraphics[width=0.8\textwidth]{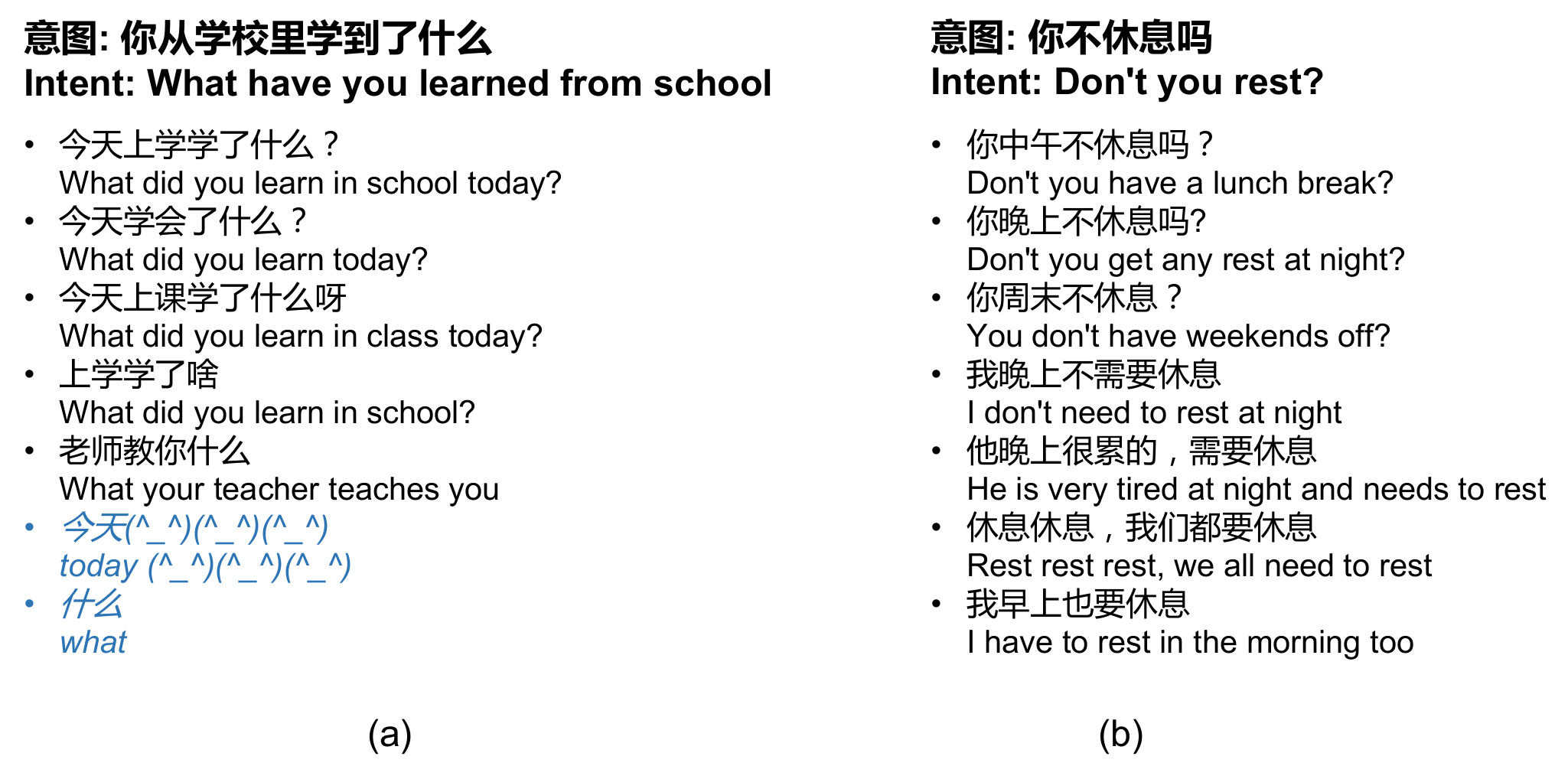}
    \caption{Mining intents from dialogue logs is an efficient way to update a task-oriented robot's intent database. We show two typical examples of real intent clusters induced from conversation logs of an MMORPG game--\textit{Ghost Story}{\footnotemark}. These two examples are clustered by the K-means algorithm, and sentence embeddings are extracted by a pre-trained bert-base-chinese{\footnotemark} model. Subfigure (a) presents an intent cluster contaminated by two outliers, which are in blue and italics.
    Subfigure (b) shows non-finetuned Bert embeddings can ensure the semantics of the sentences within a cluster are roughly similar but may fail to distinguish some subtle differences, such as negation, personal pronouns, and descriptions of time.}
    \label{fig:DDEC_intent_cluster}
\end{figure*}

\section{Related works}
\addtocounter{footnote}{-1}\footnotetext{https://qnm.163.com}
\addtocounter{footnote}{+1}\footnotetext{https://huggingface.co/bert-base-chinese}
\label{sec:print}
\textbf{Dialog Intent Induction}:
Several recent works propose to exploit the information in labeled data by combining deep models with specific loss functions or training paradigms. Perkins et al.~\cite{perkins2019dialog} propose a method exploiting multi-view data to learn representation and cluster jointly. The representations are updated iteratively using the K-means cluster assignments from the alternative view. Two other works use the self-training approach, and both incorporate K-means into the training process. The first~\cite{wang2016semi} adopted the DEC method~\cite{xie2016unsupervised}, and the second~\cite{hadifar2019self} designed a loss function that combines clustering and classification. Different from previous works, Lin et al. \cite{lin2020discovering} suggested learning a model that predicts pair-wise similarities from labeled data. The model transforms label and unlabeled data into pair-wise constraints served as the surrogate for clustering. A finetuning stage follows, and clustering is optimized based on KLD loss~\cite{xie2016unsupervised} and a process of eliminating low confidence similarity pairs is executed subsequently. However, none of the previous works considers the existence of outliers, which may make it difficult to estimate the hyperparameter $K$ of K-means. Though there exists an ITER-DBSCAN algorithm~\cite{chatterjee2020intent} that considers the outliers and claims to handle the class imbalance issue well, it does not unify representation learning and clustering algorithms under one framework. Our work remedies the above-mentioned deficiencies.\\
\textbf{Clustering Ensemble}:
Many successful applications of clustering ensemble exist in fields like image processing, cheminformatics, etc.~\cite{boongoen2018cluster}. However, applying the clustering ensemble in dialog intent induction is rare; the most relevant work is~\cite{fraj2019ensemble}. It trains base clustering models with multi-views of text representations rather than different splits of data.

\section{Density-based Deep Clustering Ensemble}
\subsection{Task Formulation}
The task is to induce novel dialog intents from unlabeled user utterances with some labeled examples in the same domain. We define the unlabeled user utterances as $D_{ul} = \{u_i, i=1,...,M\} $ and the labeled examples as $D_{l} = \{(x_{j}, y_{j}), j=1,...,N\}$, where $u_i$ denotes a user utterance, $x_{j}$ denotes the example and $y_{j} \in Y$ denotes its label. $Y$ is the collection of predefined dialog intents. $M$ is the size of the unlabeled user utterances to be processed, and $N$ is the size of the labeled examples. Given the unlabeled user utterances $D_{ul}$ and labeled examples $D_{l}$, our goal is to find a set of clusters from $T_{ul}=\{t_{i}, i=1,...,M\}$, where $t_i$ denotes the cluster label. Clusters with a size smaller than $S$ are regarded as outliers and ignored. Other clusters will be further processed, merged with existing dialog intent examples, or summarized as a novel dialog intent.
 
\subsection{Method Description}
The DDCE method consists of two steps. First, we train several base clustering models over labeled examples $D_l$. The training consists of two aspects: finetuning the text encoder and searching for the best hyperparameters. Specifically, we split, the $D_l$ into $D_l^{rl}$ and $D_l^{hs}$. $D_l^{rl}$ is the training set for finetuning the text encoder and $D_l^{hs}$ is the validation set for searching the best hyperparameters, with outliers injected. To better generalize to unseen dialog intents, we intentionally make the intents corresponding to instances in $D_l^{rl}$ and $D_l^{hs}$ not overlap each other at splitting, i.e.,  for $\forall (x_i, y_i) \in D_l^{rl}$ and $ \forall (x_j, y_j) \in D_l^{hs}$, $y_i \neq y_j$. To achieve this effect, we split $D_l$ by the dialog intents. Given split ratio $\alpha$, the collection of predefined intents $Y$, assume the size of $Y$ is $O$, we let the examples corresponding to $\alpha O$ intents enter the validation set $D_l^{hs}$, the rest of examples corresponding to other $(1-\alpha)O$ intents enter the training set $D_l^{rl}$. There are many ways to split $D_l$, from which we randomly choose $K$. For each $k\text{-th}$ split, we finetune the text encoder on $D_l^{rl}$ with a classification task and the cross-entropy objective function. We split another validation set from $D_l^{rl}$ to select the best model according to the classification performance. Given the best text encoder, we conduct a random search of hyperparameters on $D_l^{hs}$ and choose the best one with the highest $score$, a metric later described in the evaluation setting. At last, we obtain $K$ base clustering models, respectively corresponding to $K$ text encoders and $K$ groups of hyperparameters. We shall calculate the performance score $score_c$ of the $K$ base clustering models on their respective corresponding validation sets $D_l^{hs}$ for the following ensemble. The details of calculating $score_c$ are covered in Evaluation Metric Section.

Second, we use $K$ base clustering models to do clustering over $D_{ul}$, obtaining $K$ groups of cluster labels, of which the $k$th labels can be noted as $T_{ul}^k$. Finally, we apply a consensus function over the results $(T_{ul}^1,...,T_{ul}^K;score_c^1,...,score_c^K)$ to obtain final clustering labels $T_{ul}$. The complete process of the algorithm is summarized in Algorithm \ref{alg:ddec}.

\subsubsection{Consensus Function}
\label{sssec:subsubhead1}
Combining the clustering results of base models is non-trivial because of the label correspondence problem~\cite{strehl2002cluster}. We introduce three consensus functions here and test BOKV and CHM in experiments. We denote $NMI(\cdot, \cdot)$ as the normalized mutual information, and $\mathbb{T}_{K} = \{T_{ul}^1,...,T_{ul}^K\}$ as the set consisting all base models' partitions.

\textbf{CSPA/HGPA/MCLA (CHM)~\cite{strehl2002cluster}}:
CHM uses three partition methods to generate different cluster labels.
The optimal labels $T_{ul}^{*}$  for CHM is defined as
\begin{equation}
T_{ul}^{*}=\underset{T \in \mathbb{T}_{chm}}{\arg \max } \;\sum_{j=1} \mathrm{NMI}\left(T, T_{j}\right)
\end{equation},
where $\mathbb{T}_{chm} = \{T_{CSPA}, T_{HGPA}, T_{MCLA}\}$. Abbreviations \textbf{CSPA}, \textbf{HGPA} and \textbf{MCLA} denote Cluster-based Similarity Partitioning, HyperGraphs Partitioning, and Meta-CLustering Algorithm respectively~\cite{strehl2002cluster}.

\textbf{Best Of K (BOK)~\cite{vega2011survey}}: In BOK, the optimal labels $T_{ul}^{*}$ is defined as 
\begin{equation}
T_{ul}^{*}=\underset{T \in \mathbb{T}_{K}}{\arg \max}\;\sum_{j=1}^{K} \mathrm{NMI}\left(T, T_{j}\right)
\end{equation}

\textbf{Best Of K with outlier Voting (BOKV, ours)}: As all base models share the same label for outliers, to boost performance, we choose to aggregate outliers' predictions via simple voting~\cite{bauer1999empirical}. Because the low performances of base models may have a negative impact on the ensemble result~\cite{wang2008some}, only when more than half of base models' non-outliers recall scores on validation set are higher than 0.5 do we vote to predict outliers, otherwise, BOKV is degraded to BOK.
When BOKV is adopted, for each sample $t_i$, we first obtain the prediction $u_i$ by majority voting, to decide whether it is outlier or non-outlier,

\begin{equation}
u_{i}=\left\{\begin{array}{ll}
1 & \text { if } \operatorname{argmax}_{t_j}\left\{\sum_{j=1}^{K} t_k \right\}=l^{out} \\
0 & \text { if } \operatorname{argmax}_{t_j}\left\{\sum_{j=1}^{K} t_k \right\} \neq l^{out}
\end{array}\right.
\end{equation},
where $l^{out}$ is the label of outlier. We denote $I_{out}=\{i\,|\,u_{i}=1\}$ and  $I_{nout}=\{i\,|\,u_{i}=0\}$ as index sets of outlier and non-outlier respectively.
While the ensemble labels of outliers $T_{ul}^{out}=\{t_{i}, i \in I_{out} \}$ are determined by voting, the ensemble labels of non-outliers $T_{ul}^{nout}$ still need to be determined by BOK,
\begin{equation}
\label{eq:t_nout}
T_{ul}^{nout}=\underset{T^{nout} \in \mathbb{T}_{K}^{nout}}{\arg \max } \;\sum_{j=1}^{K} \mathrm{NMI}\left(T^{nout}, T_{j}^{nout}\right)
\end{equation}
where $T^{nout} = \{t_i, i \in I_{nout}\}$ denotes the partition of a base learner with non-outliers. At last, we obtain the optimal label $T_{ul}^{*}$ by combining $T_{ul}^{out}$ and $T_{ul}^{nout}$.

\begin{algorithm}[h]
\SetAlgoLined

 \textbf{Input}: $D_{l} = \{(x_{j}, y_{j}), j=1,...,N\}$, \\
  \hphantom{\textbf{Input}:} $D_{ul} = \{u_i, i=1,...,M\}$\;
  \textbf{Ouput}: $T_{ul}$ = \{$t_{i}, i=1,...,M\}$\;
   
 \textbf{Require}: The number of base cluster models $K$, the collection of predefined intents $Y$, the size of $Y$ is $O$, the split ratio $\alpha$, hyperparameter search space ${hp}$\;
 
 \textbf{Training}:\\
 \For{$k = 1,...,K$}{
    split $D_{l}$ into $D_{l}^{rl}$ which contains examples corresponding to $(1-\alpha)O$ intents, and $D_{l}^{hs}$ which contains examples corresponding to $\alpha O$ intents \\
    Initialize text encoder  $f_{\theta}^{k}$ with pre-trained weights\\
    Update $\theta^{k}$ after trained on $D_{l}^{rl}$ \\
    Compute embeddings $E_{l}^{hs}$ of $D_l^{hs}$ using $f_{\theta}^{k}$ \\
    Search the best hyperparameters $hp^{k}$ on $E_{l}^{hs}$ \\
    Calculate $score_{c}^{k}$ on $D_{l}^{hs}$ with $hp^{k}$
  }
  \textbf{Inference}:\\
  \For{$k = 1,...,K$}{
  Compute embeddings $E_{ul}$ of $D_{ul}$ using $f_{\theta}^{k}$ \\
  Do clustering over $E_{ul}$ with hyperparameters $hp^k$
  }
  Apply a consensus function (e.g. BOKV) on $(T_{ul}^1,...,T_{ul}^K;score_c^1,...,score_c^K)$ to obtain $T_{ul}$
 \caption{DDCE}
 \label{alg:ddec}
\end{algorithm}

\section{Experiments}
\label{sec:Experiments}

\subsection{Datasets and Preprocessing}
\begin{table}[ht]
\centering
\setlength{\tabcolsep}{2mm}
\begin{tabularx}{\linewidth}{lcccc}
\toprule
Dataset &  Class &   Text &  Len &  U.Token \\
\midrule
CLINC150 (EN) &     150 &   150 &   40 &     6391 \\
DBpedia (EN) &     219 &  1566 &  121 &   418737 \\
IWSDS (EN)    &      68 &   377 &   35 &    10585 \\
THUCNews (CN)    &      14 &  59720 &   20 &   266060 \\
SMP2019 (CN)    &      23 &    156 &    9 &     3482 \\
AgentDialog (CN) &     354 &     21 &    7 &     2883 \\
HumanDialog (CN) &    1226 &     15 &    6 &     2790 \\
\bottomrule
\end{tabularx}
  \caption{The columns from left to right show the number of classes, the average number of text in each class, the average length of the text, and the total number of unique tokens.}
  \label{tab:1}
\end{table}

We present the statistics for datasets used in the experiments in Table 1, including three English datasets and three Chinese datasets. CLINC150~\cite{larson-etal-2019-evaluation} is an intent classification dataset with 150 in-domain intent classes. DBpedia~\cite{auer2007dbpedia} is an active project dealing with structured data and Wikipedia. IWSDS~\cite{XLiu.etal:IWSDS2019} is a multi-domain SLU benchmarking dataset built from 25K user utterances. THUCNews is a news classification dataset\footnote{http://thuctc.thunlp.org}. SMP2019 is the dataset of SMP2019 ECDT Task1\footnote{https://conference.cipsc.org.cn/smp2019/evaluation.html}. AgentDialog and HumanDialog are user utterances respectively extracted from agent-human and human-human conversation logs of an MMORPG game--\textit{Ghost Story}\footnote{https://qnm.163.com}, where agents are intelligent kids. To make the experiments closer to the real-life scenario, we inject outliers into the test sets. The outliers are samples from other datasets. For example, we may randomly pick one sample per intent from DBpedia and add them to the test set of CLINC150 as outliers. Ratios of injected outliers in experiments are presented in Table \ref{tab:2}.

\subsection{Evaluation Metrics}
\label{ssec:eval_metrics}
To measure the performance of novel intents detection, we considered two metrics: $score_{c}$ to measure the recall of non-outlier samples, following Lin etc.~\cite{lin2020discovering}, we adopt the Adjusted Rand Index score~(ARI)~\cite{yeung2001details} to measure the clustering quality of $T_{ul}$, denoted as $score_{ari}$. The final ${score}$ is defined as the harmonic mean of $score_{c}$ and $score_{ari}$. We use the harmonic mean because we believe that the ability to detect outliers and the clustering quality are equally important. We hope as many reasonable clusters to be found as possible, at the same time, the quality of which is good enough to be readily merged with old intents or form as new ones with less or no post-processing. In practice, we found numerous outliers during the process of inducing new intents from human-human conversation logs in online games, thus we believe the new metric is more in line with the real world.

\subsection{Baselines}
\label{ssec:subhead4}
We include unsupervised clustering algorithms --- K-means~\cite{steinley2006k}, Hierarchical Clustering using Ward Linkage~\cite{murtagh2011ward} and OPTICS~\cite{ankerst1999optics}. We also compare with the highly influential DEC~\cite{xie2016unsupervised}, a method that proposes an iterative refinement via soft assignment. For semi-supervised clustering, we compare with SOTA works including BERT-MCL~\cite{hsu2018multi} and CDAC+~\cite{lin2020discovering}. We use BERT-base as the text encoder in every method to ensure a fair comparison. All text encoders are in-domain fine-tuned except for the OPTICS baseline.

\begin{table*}[t]
\centering
\setlength{\tabcolsep}{1.5mm}
\resizebox{\textwidth}{23mm}{
\begin{tabular}{cccccccc}
\toprule
&   CLINC150 &          DBpedia &    IWSDS & THUCNews &          SMP2019 &      AgentDialog &      HumanDialog \\
\midrule
      Outlier ratio            &        0.547          &       0.279           &         1.143         &       2.0           &           2.0       &      1.144      &    0.156    \\
\midrule
K-means      &  0.208$\pm$0.031 &  0.144$\pm$0.009 &  0.128$\pm$0.004 &  0.458$\pm$0.083 &  0.365$\pm$0.064 &  0.305$\pm$0.014 &  0.605$\pm$0.012 \\
Hierarchical &  0.248$\pm$0.043 &  0.169$\pm$0.016 &  0.146$\pm$0.021 &  \underline{0.504$\pm$0.027} &  0.521$\pm$0.126 &  0.356$\pm$0.018 &  0.622$\pm$0.003 \\
DEC          &  0.102$\pm$0.034 &  0.089$\pm$0.011 &  0.125$\pm$0.045 &  \textbf{0.533$\pm$0.135} &  0.427$\pm$0.042 &  0.288$\pm$0.054 &  0.683$\pm$0.016 \\
BERT-MCL          &  0.307$\pm$0.237 &  0.025$\pm$0.030 &  0.036$\pm$0.033 &  0.046$\pm$0.079 &  0.231$\pm$0.204 &  0.200$\pm$0.104 &  0.015$\pm$0.017 \\
CDAC+        &  0.203$\pm$0.002 &  0.287$\pm$0.044 &  0.119$\pm$0.044 &  0.007$\pm$0.013 &  0.118$\pm$0.046 &  0.360$\pm$0.042 &  0.585$\pm$0.011 \\
OPTICS       &  0.247$\pm$0.079 &  0.185$\pm$0.063 &  0.199$\pm$0.111 &  0.375$\pm$0.176 &  0.487$\pm$0.195 &  0.348$\pm$0.122 &  0.759$\pm$0.058 \\
DDEC-CHM (ours)    &  \textbf{0.563$\pm$0.046} &  \underline{0.315$\pm$0.035} &  \textbf{0.517$\pm$0.115} &  0.263$\pm$0.241 &  0.408$\pm$0.274 &  0.515$\pm$0.059 &  0.744$\pm$0.044 \\
DDEC-BOKV-BM (ours) &  0.525$\pm$0.008 &  0.300$\pm$0.023 &  0.411$\pm$0.017 &  0.469$\pm$0.077 &  \underline{0.700$\pm$0.067} &  \underline{0.605$\pm$0.015} &  \underline{0.853$\pm$0.011} \\
DDEC-BOKV (ours)   &  \underline{0.557$\pm$0.057} &  \textbf{0.363$\pm$0.021} &  \underline{0.506$\pm$0.058} &  \underline{0.504$\pm$0.140} &  \textbf{0.760$\pm$0.080} &  \textbf{0.641$\pm$0.010} &  \textbf{0.855$\pm$0.015} \\
\bottomrule
\end{tabular}}
  \caption {The values in the table correspond to the $score$ described in the Evaluation Metrics Section. The \emph{outlier ratio} is the ratio of the size of injected outliers to the original size of $D_{ul}$. \textbf{BOKV-BM} refers to the average performance of $K$ BOKV's base models.
}
  \label{tab:2}
\end{table*}

\subsection{Experimental Settings}
\label{sec:exp_set}
The experiment is repeated three times for each dataset, with different splits of $D_l$ and $D_{ul}$. $D_{ul}$ includes examples corresponding to around 15\% of total dialog intents before outliers injection. 

To approximate the situation in real applications where there are only a small number of labeled examples per intent, the maximum size of samples per intent is set to 50. 

We set the number of base clustering models $K$ to 5 and split ratio $\alpha$ to 0.5 throughout all experiments. For both languages, we chose BERT-base as our text encoder because BERT is the basis for most SOTA text encoders\footnote{https://gluebenchmark.com/leaderboard}. We set the batch size to 32 and the learning rate to $5e^{\\-5}$. The embedding extracted for clustering is the average pooling of the second last layer of BERT-base. 

Regarding the hyperparameter searching phase, we chose OPTICS~\cite{ankerst1999optics} as the density-based clustering algorithm; compared to the popular algorithm DBSCAN, it has the advantage of finding clusters with varying densities. We conducted random searches~\cite{bergstra2012random} on three hyperparameters of OPTICS: ${max\ eps}$, $xi$, and $min\ sample$. The interval of $min\ sample$ is set to $(2, 20)$, and the intervals of ${max\ eps}$, $xi$ are set to ${(0.0, 0.5)}$ respectively. We repeat the search process a hundred times per trial.

For K-means based methods, which require setting the number of clusters $K_c$, we first estimate the average size of examples per intent in the labeled data $D_l$ and infer the $K_{c}$ for test set $D_{ul}$ accordingly. To boost baseline methods' performance, we increase the $K_{c}$ by a factor of 4 for a rough estimate of outliers. We leave any example in clusters with a size smaller than 2 as outliers.

\subsection{Results}
\label{sec:res}
The results on seven datasets are shown in Table \ref{tab:2}. We tested our method \textbf{DDEC} with two consensus functions: CHM and BOKV. Concerning performance, DDEC-CHM and DDEC-BOKV outperform all other methods on all datasets except for the THUCNews dataset, where DEC ranks first. However, we find DEC quite sensitive to the choice of degree of freedom; in Table~\ref{tab:2}, the reported $scores$ of DEC are the best ones chosen from trails of different hyperparameters. From the above observations, we conclude that K-means related methods are more susceptible to inappropriate hyperparameters while the \textbf{DDEC} is more robust and ranks at the top for all datasets.

The overall best consensus function is BOKV, though it slightly underperforms CHM on CLINC150 and DBpedia. The ensemble method DDEC-BOKV exhibits a consistent advantage over its base models in all datasets, demonstrating better generalization in discovering new intents.\\

\begin{figure}[h]
    \centering
\includegraphics[width=0.45\textwidth]{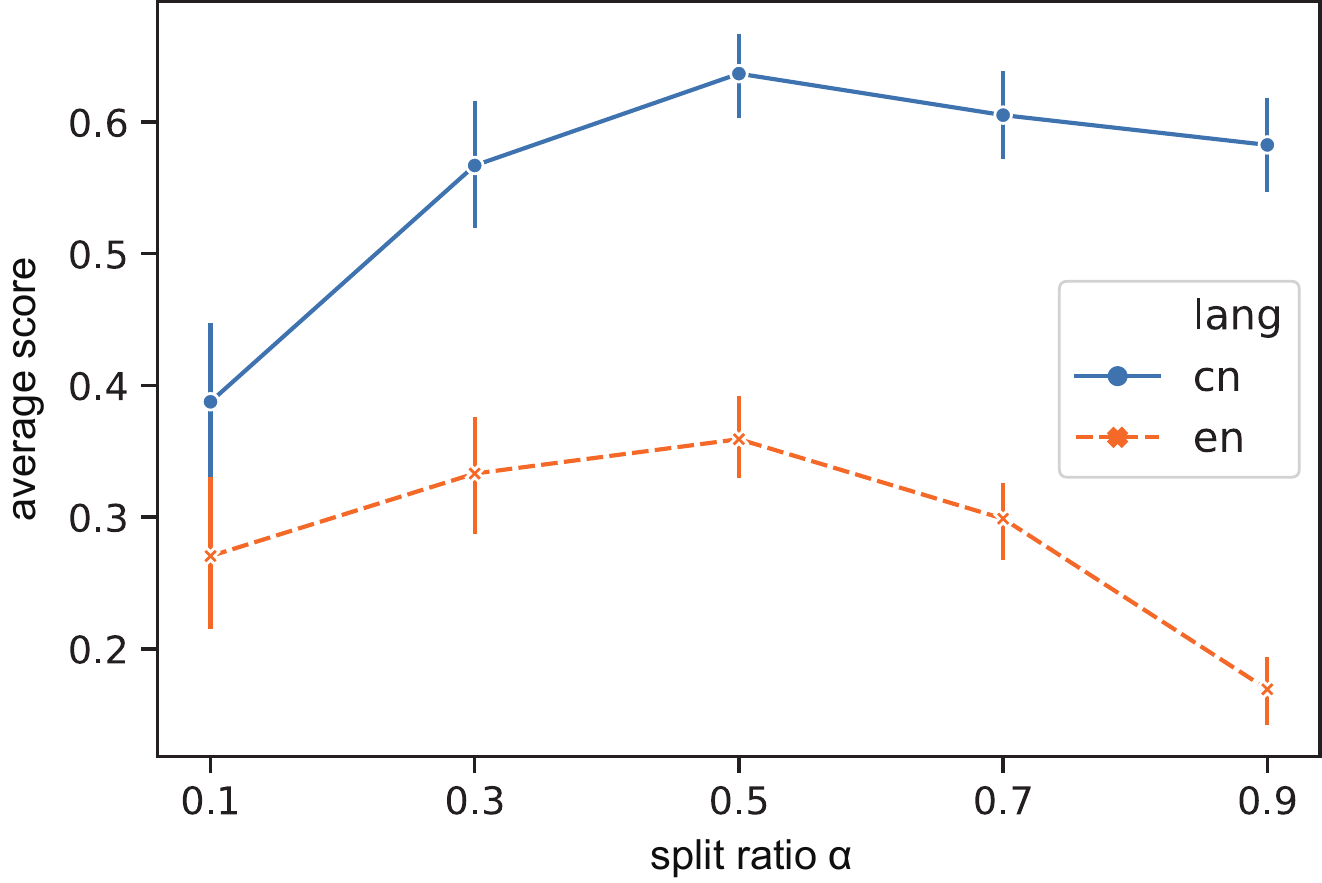}
    \caption{The impact of split ratio $\alpha$ for a base clustering model.}
    \label{figure:ratio_to_score}
\end{figure}

\textbf{The best split ratio $\alpha$}:
We test how the choice of $\alpha$ impacts the performance of a base clustering model. For each $\alpha$ value, we repeat the same experimental procedure described in the Experimental Section. For both English and Chinese datasets, scores are averaged over datasets. In Figure~\ref{figure:ratio_to_score}, we show how the mean and variance of the scores are influenced by $\alpha$. We can conclude that setting $\alpha$ at 0.5 achieve an optimal balance regardless of the language of the dataset.

\begin{figure}[H]
    \centering
\includegraphics[width=0.45\textwidth]{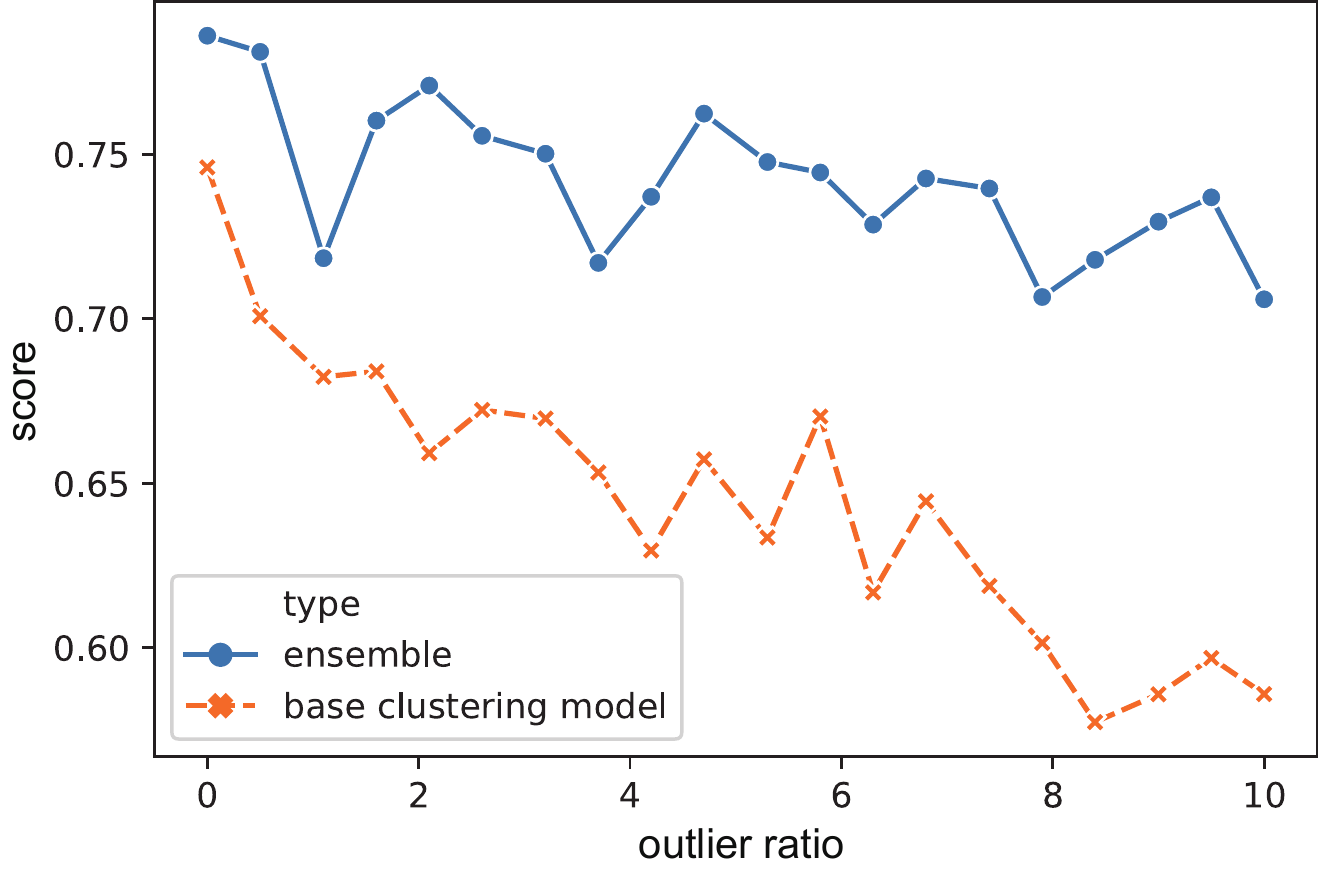}
    \caption{We present the effect of outliers ratios on the results. We choose SMP2019 as the experimental dataset to test a larger order of magnitude of outlier ratios.}
    \label{figure:outlier_ratio_to_score}
\end{figure}

\textbf{Effects of different ratios of outliers}:
Trends in Figure \ref{figure:outlier_ratio_to_score} show that the score of DDEC-BOKV decreases much more slowly than the average score of base clustering models as the outlier ratio increases, indicating that BOKV introduces more robustness against the number of outliers.

\textbf{Effects of different sizes of training data}:
We conduct experiments to analyze if DDEC-BOKV is sensitive to the labeled data's size $D_l$. Table~\ref{tab:3} proves convincingly that the clustering ensemble consistently outperforms base clustering models, except the difference is only significant at $p=0.07$ when $O$ is minimal. This experiment demonstrates that BOKV significantly outperforms the base model even when the data volume is very small.
\begin{table}[H]
\centering
\setlength{\tabcolsep}{1.5mm}
\begin{tabularx}{\linewidth}{cccccccc}
\toprule
 $O$ & 4   &    8   &    16  &    32  &    64  &    128 \\
\midrule
  & 10.4\% &  15.2\%\dag &  9.9\%\dag &  10.0\%\dag &  4.2\%\dag &  6.3\%\dag \\
\bottomrule
\end{tabularx}
  \caption{We report the relative $score$ improvement of DDEC-BOKV over DDEC-BOKV-BM when training size varies. $O$ denotes the size of the collection of predefined intents $Y$ in $D_l$. Significant tests are performed~\cite{woolson2007wilcoxon} and $\dag$ indicates ${p < 0.05}$.}
  \label{tab:3}
\end{table}

\section{Conclusion}
\label{sec:subhead8}
In practice, we find that clusters mined in conversation logs by K-means based clustering algorithms often contain many outliers, partly because of the characteristics of the data itself, and partly because K-mean based clustering algorithms alone cannot handle outliers properly. To compensate for the shortcomings of the K-means based methods, we propose a deep clustering ensemble method as well as a new outlier-aware metric for the dialog intent induction task. Our approach encourages base models to learn from different parts of the labeled data. We maximize the use of data through finetuning a text encoder and searching a proper set of hyperparameters for OPTICS simultaneously. To avoid overfitting, separate clustering results are integrated via a novel consensus function BOKV. Our method is proved effective in extensive experiments, even if the size of labeled data is extremely small or the unlabeled data contains a large number of outliers.

\bibliography{main}
\end{document}